\documentclass[fleqn,10pt]{olplainarticle}
\usepackage{graphicx}
\usepackage{float}
\usepackage[]{subfig}
\usepackage{multirow}
\usepackage{amsmath}
\usepackage{verbatim}
\usepackage{pdfpages}
\usepackage{wrapfig}

\usepackage{flushend}

\title{Leader-Follower Dynamics in Complex Obstacle Avoidance Task}

\author[1,2]{Rebeka Kropivšek Leskovar}
\author[1,3]{Jernej Čamernik}
\author[1,2]{Tadej Petrič}

\affil[1]{Department for Automation, Biocybernetics and Robotics,Jo\v{z}ef Stefan Institute, Jamova cesta 39, Ljubljana, Slovenia}
\affil[2]{Jo\v{z}ef Stefan International Postgraduate School, Jamova cesta 39, Ljubljana, Slovenia}
\affil[3]{Applied Cognitive Psychology, Ulm University, Helmholtzstraße 16, Ulm, Germany}


\begin{abstract}
A question that many researchers in social robotics are addressing is how to create more human-like behaviour in robots to make the collaboration between a human and a robot more intuitive to the human partner. In order to develop a human-like collaborative robotic system, however, human collaboration must first be better understood. Human collaboration is something we are all familiar with, however not that much is known about it from a kinematic standpoint. One dynamic that hasn't been researched thoroughly, yet naturally occurs in human collaboration, is for instance leader-follower dynamics. In our previous study, we tackled the question of leader-follower role allocation in human dyads during a collaborative reaching task, where the results implied that the subjects who performed higher in the individual experiment would naturally assume the role of a leader when in physical collaboration. In this study, we build upon the leader-follower role allocation study in human dyads by observing how the leader-follower dynamics change when the collaborative task becomes more complex. Here, the study was performed on a reaching task, where one subject in a dyad was faced with an additional task of obstacle avoidance when performing a 2D reaching task, while their partner was not aware of the obstacle. We have found that subjects change their roles throughout the task in order to complete it successfully, however looking at the overall task leader the higher-performing individual will always dominate over the lower-performing one, regardless of whether they are aware of the additional task of obstacle avoidance or not.
\end{abstract}

\begin{document}

\flushbottom
\maketitle
\thispagestyle{empty}

\section{Introduction}\label{sec1}
As more and more robots work in direct contact with people, finding the answers to the question of effective human-robot collaboration (HRC) is becoming imperative. Due to this, a great amount of state-of-art research has been done on the subject of HRC over the past decade. One of the main issues being how to develop robots that will be accepted by their human partners \cite{Brohl2019}. Studies such as \cite{Noohi2016} showed that human-like behaviour in robots has a positive effect on the human perception of their robot partner, as well as on the task performance. They have found that when collaborating with a robot, people find human-like behaviour more intuitive. Moreover, a study by Ivanova et al. \cite{Ivanova2020} found that humans even prefer a robot partner with human-like behaviour to an actual human partner for motion assistance as, when designed correctly, they are perceived to be more predictable than human partners. 

In this regard, many human-based control models have already been developed, such as \cite{Petric2016,Leica2016,Petric2017,Khoramshahi2018,Li2020}. Although trying to imitate human behaviour \textit{in} a robot partner, these studies do not however extend their imitation to the behaviour \textit{between} the robot and the human partner. In order to implement dynamics between partners into HRC systems, we must first however better understand how such behaviours naturally occur in human collaboration.

Human collaboration is something that we all face on a daily basis. Whether it's when working on a group project at work or when moving furniture at home. One of the key aspects of successful human collaboration is communication. Cohen’s Joint Intention Theory, presented in \cite{Cohen1991}, states that performing a collaborative task efficiently in a changing environment requires an open channel of communication. This allows the partners to coordinate between themselves and adjust their behaviour in real-time in order to reach a common goal. However, this is not limited to only verbal communication but non-verbal forms of communication as well; more specifically through the use of physical interactions. 

For instance, a study by Ganesh et al. \cite{Ganesh2014} found that the partners were able to improve their task performance when collaborating with each other without verbal communication or even knowing that they were collaborating. This shows that physical coordination is crucial to completing a task successfully. The study also showed evidence that the task performance of both partners improved, even when the partner of the subject was inferior in performing the task. The mutual improvement during continuous interaction was explained in a paper by Takagi et al. \cite{Takagi2017} as a result of a mechanism where individuals are able to estimate the partner’s target from the interaction force to improve their prediction of the target’s motion. The results from \cite{Takagi2017} were further examined in a later study by Takagi et al. \cite{Takagi2018} in which they showed that the strength of the connection between partners also has an influence on the task performance, where a stronger connection yields a better estimate of the partner’s target, enabling partners to improve more from the interaction. In a similar vein, a study by Batson et al. \cite{Batson2020} showed that the experiment group, where participants were coupled together, outperformed the control group in solo trials, meaning collaborating with a partner also improved future solo performance. It should be noted, however, that no real consensus has been found so far on whether coupled partners always outperform solo performance, as results from studies by Che et al. \cite{Che2016} and Beckers et al. \cite{Beckers2018} seem to contradict the findings in \cite{Ganesh2014}.

However, communication is not the only aspect that influences how successful collaboration is. Another key aspect of successful human collaboration is also role allocation. In a review by Sebanz et al. \cite{Sebanz2006} the authors showed that role allocation is one of the important areas in human collaboration that hasn't yet been sufficiently researched. Furthermore, the importance of role distributions in human-robot collaboration has been addressed in \cite{Jarrasse2014}. Since the publication of the review by Sebanz et al., more research in this area has been done. For instance, in the work by van der Well et al. \cite{Wel2011} an indication that haptic information may be the key component to solve the leader-follower role allocation problem can be found. Furthermore, the existence of an unequal control in human collaboration was observed in \cite{groten2009experimental}, where they focused on the influence of haptic feedback on dominance distribution. Here they found that participants work with a consistent dominance distribution, independent of the different feedback conditions. In \cite{stefanov2009role} they even proposed that one partner is responsible for the plan and the other is responsible for executing the actions. On the other hand, in \cite{Takagi2016} researchers found that rigidly coupled pairs perform tasks based on pre-programmed motion plans, independent of their partner's behaviour. Role allocation was more recently investigated in \cite{sheybani2020evolving}, where they used genetic algorithms to evolve simulated agents which explored archetypes of cooperative role switching strategies in humans. They found that dyads with anti-synchrony, i.e. dyads where agents assumed complementary roles, performed the best. However, human experiments were not performed in order to confirm the simulation results. 

In our previous study \cite{Kropivsek2021}, we tackled the question of leader-follower role allocation in human dyads during a collaborative reaching task. We examined how subjects in collaboration influence the task performance and whether both partners' influence over the task is equal or if one partner seems more dominant over the other, thus establishing a leader-follower dynamic within a dyad. Furthermore, we have investigated whether the prominence of one partner in a collaborative task can be predicted by observing the partner's individual task performance. The results provided in the study implied that the subjects who performed higher in the individual experiment would naturally assume the role of a leader when in physical collaboration with another subject. However, the experiments were only performed on a simple 2D reaching task. To examine whether the findings from this study hold even when faced with different tasks or if role allocation varies based on the task at hand, a similar experiment with a different, more complex task needs to be conducted.

In this study, we continue examining the leader-follower role allocation in human dyads by observing how the leader-follower dynamics change when one subject in a dyad is faced with an additional task of obstacle avoidance when performing a 2D reaching task, while their partner is not aware of the obstacle. Here we hypothesise that when subjects see an obstacle, in order for them to avoid it successfully, they must assume the role of the leader regardless of their primary role allocation, which is designated based on the results from our previous study \cite{Kropivsek2021}. Furthermore, we hypothesise that although the subjects who see the obstacle will be the leader of the dyad, the subjects who would naturally assume the role of the leader will have an easier time completing the experiment successfully. Meaning, when a subject who is predicted to be the leader based on our previous study is aware of the obstacle, the number of failed trials will be lower compared to when a subject who is predicted to be the follower is the only one aware of the obstacle.

\section{Methods}\label{sec:methods}
\subsection{Subjects}
The study included twelve male and four female subjects, with an average height of 180.25 cm and 167.5 cm respectively. All subjects were right-handed. The average age of the subjects was 28.3 years old. Prior to their participation, the subjects were informed about the experimental procedure, potential risks, the aim of the study and gave their written informed consent in accordance with the code for ethical conduct in research at Jožef Stefan Institute (JSI). This study was approved by the National Medical Ethics Committee (No.: 0120-228/2020-3).

\subsection{Experimental Setup}

\begin{figure}[t]
    \subfloat[]{\includegraphics[clip,trim=0cm 0cm 0cm 4cm,width=0.5\textwidth]{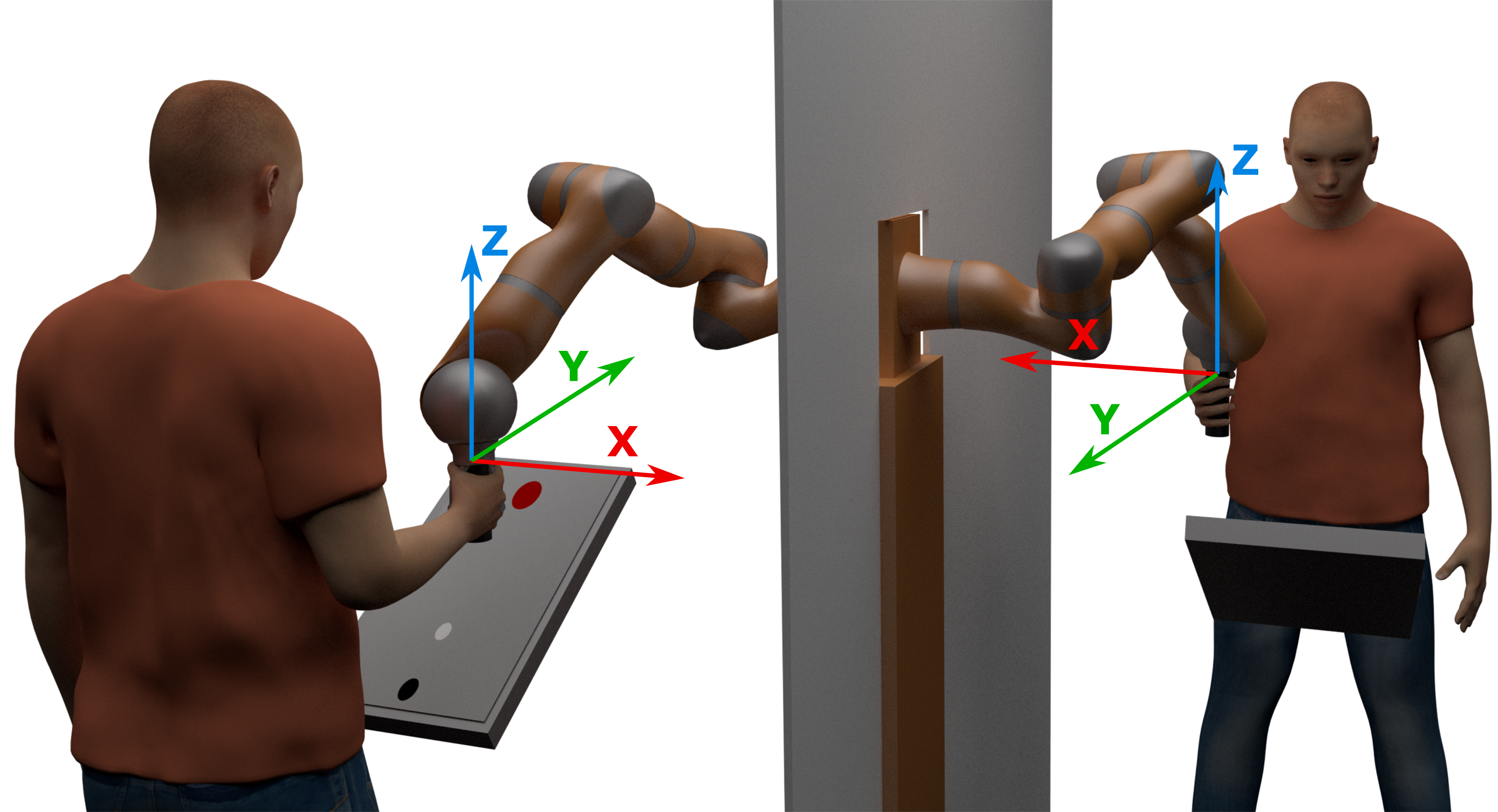}}
    \hspace{0.5cm}
    \subfloat[]{\includegraphics[clip,trim=1cm 1cm 1cm 2cm,width=.5\textwidth]{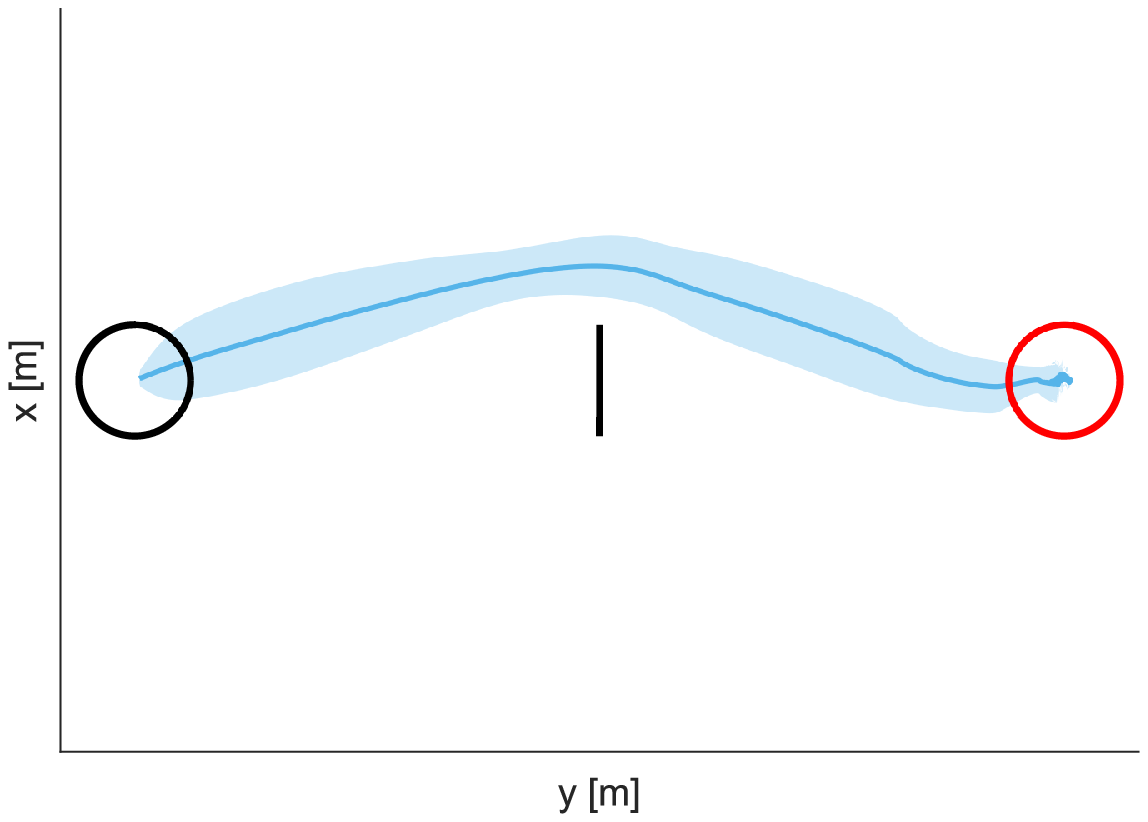}}
	\caption{a) Experimental setup. b) Sketch example of the experimental task. The sketch shows the starting point (black circle), target point (red circle), the obstacle (black line) and an average trajectory of the subjects in collaboration with standard deviation.
	}\label{fig:experiment}
\vspace*{-0.4cm}
\end{figure}

The experiments were conducted on a dual-arm Kuka LWR robot, seen in Fig. \ref{fig:experiment}. The robots were used as a haptic interface between the subjects and the virtual environment. Here, the robot arms were used as two separate haptic interfaces for individual subjects or as one combined haptic interface, where the robot arms and their users were coupled together through a virtual dynamic model. This was achieved using a virtual dynamic model for the robot arms developed in \cite{Leskovar2020}.  

The virtual dynamic model consists of two points, each representing one end-effector of the dual-arm robot and a controlled point, situated between the two end-effector points, always at an equal distance from the two. When a  force is applied to any of the two end-effectors, the dynamic model generates a proportional force to the controlled point. This proportional force is the sum of the forces applied to both end-effectors when the subjects are coupled together. Due to this system, when coupled together the two subjects move the same virtual point in unison by applying their individual force to each end-effector, which establishes an open channel of communication through haptic interaction, similar to the haptic communication channel seen in \cite{Wel2011}. On the other hand, when the subjects are not coupled together, each subject controls a separate virtual point, where the proportional force now equals the force applied to the subject's end-effector.

As a visual interface two monitors, displaying a graphic user interface (GUI) shown in Fig. \ref{fig:experiment}, were used in the experiment. Here, the red dot represents the target, whose position and size change throughout the experiment. The black dot represents the starting position and the white dot the controlled point. To match the GUI, the movement of the robot arms was limited to a 2D plane. This was done by constraining the z-axis of the robot's end-effector (see \ref{fig:experiment}) to a static position so that the angle between the subject's arm and forearm was 90deg in the starting position. 

\subsection{Experimental Protocol}

The experiment consisted of two sets, where in the first set the subjects performed the required task individually with no obstacles present, while in the second set the subjects were coupled together to perform the task in collaboration. Moreover, in the second experimental virtual obstacles were included in the task. However, only one subject was made aware of them.

Each experimental set began with the subjects moving the controlled point on the screen to its starting position. When the controlled point was in its starting position, a random target appeared on the screen. The subjects were instructed to reach this target and stay inside until the target disappeared. This was to prevent the subjects from simply running over the target without aiming for it. When the target disappeared, the subjects had to return to the starting position. The reaching task was repeated 90 times in each set in which 9 different targets with varying distances (5cm, 15cm and 25cm) and size (small, medium and large) were used in random order.

In the second experimental set, the subjects were coupled together in order to collaborate. Before the experimental set started one of the subjects was designated to get an additional task of obstacle avoidance, while their partner was unaware of the obstacles. The subjects were not made aware in advance which partner received which role. Which subject saw the obstacle depended on their performance in the individual experimental set. In half of the dyads, the subject with lower individual performance was selected as the one to see the obstacle, while in the other half of the dyads the subject with higher individual performance was selected to be the one who sees the obstacle. 

The collaborative experimental set was conducted in the same manner as the individual experimental set with the addition of obstacle avoidance. This meant that one subject in a dyad had only the task of reaching a target, while their partner had the task of reaching a target whilst avoiding an obstacle. The obstacle appeared on the screen at the same time as the target and was always positioned in the centre of the screen, midway between the starting position and the target. Meaning, when the target distance was 5 cm, the obstacle appeared on the screen at a 2.5 cm distance from the starting position. If the subject who was aware of the obstacle was unsuccessful in avoiding it, they failed the task and had to return to the starting position without reaching the target. At the same time, the partners were decoupled so that the subject who was not aware of the obstacle could finish their reaching task. Furthermore, the subject who was not aware of the obstacle was not informed that the other subject failed.

After each experimental set, the subjects had a 5-minute rest during which they were asked to fill out the NASA Task Load Index questionnaire.

Throughout the experiment, the subjects were not told whether they were performing a collaborative or an individual task. However, when the two subjects were coupled together, they could feel an external force produced by their partner. This established an open channel for haptic communication between the two subjects, which allowed them to sense when they were performing the task together or alone.

\subsection{Data Processing}\label{sec:dataprocessing}
\textit{\textbf{Determining higher-performing/lower-performing subject:}}
The gathered data from the experiment was divided into two tasks - the collaborative task and the individual task. The data sets from the individual task were then further divided based on subjects performance during individual trials. Here, subjects were determined to be either the individual with the higher performance or the individual with the lower performance. This was done by comparing the subjects' average maximum velocity in individual tasks, where the subject with a higher maximum velocity is defined as the higher-performing individual. The used sorting method was decided upon from the results of our previous study \cite{Kropivsek2021}, where we found that observing the maximum velocity in individual tasks produces the most accurate prediction of the leading subject in the collaborative task.

\textit{\textbf{Determining the leader-follower dynamics:}} The leader of the collaborating task was determined in the same manner as in our previous study \cite{Kropivsek2021} -- by measuring and analysing the difference in the forces applied by the partners in collaboration ($\Delta \boldsymbol{F}$):

\begin{equation}\label{eq:dF}
    \Delta\boldsymbol{F} = \lvert\boldsymbol{F}_{hp}\rvert - \lvert\boldsymbol{F}_{lp}\rvert 
\end{equation}

where $F_{lp}$ is the force applied by the partner with lower performance as an individual and $F_{hp}$ is the force applied by the partner with higher performance as an individual. The overall leader of each trial was then defined as
\begin{equation}\label{eq:leader}
    L = \int_{0}^{T} \Delta\boldsymbol{F}(t) dt
\end{equation}
where $T$ is the measured time it took to reach the target. To determine the overall leader for each target an average leader value $L$ is calculated from all trials. 

The difference in forces was used for determining the leader since the robot movement is determined solely by the force applied to its end-effector. Meaning the subject who applied a higher force to the end-effector had a higher influence on the robot movement, which in turn allocated them as the leader of the task at hand.

\textit{\textbf{Determining task performance:}} The task performance of individual and collaborative experiments was additionally evaluated to observe whether the task performance improved when subjects were in collaboration, as was shown in \cite{Ganesh2014}, despite having an additional task of avoiding an obstacle. This was done by taking into account the average Fitts' law's Index of Performance (IP), described as: 

\begin{equation}\label{eq:ip}
    IP = \frac{ID}{MT} ,
\end{equation}
where $MT$ is the measured movement time and $ID$ is the index of difficulty,
which has several formats in literature as seen in \cite{Fitts1954},\cite{Zhai2004}. 

In this study the Shannon formulation \cite{MacKenzie1992} was used, which is defined as:
\begin{equation}\label{eq:id}
    ID = log_2(\frac{D}{W} + 1) .
\end{equation}
Here, $D$ is the distance of the target and $W$ is the width of the target. 

The IP was calculated for all targets in each trial. The calculated IPs were then averaged within each target. The overall IP for each subject was then determined by calculating the average IP from all targets. When comparing the IPs of both individuals, the subject with a higher IP is determined to be the subject with superior individual performance. 

Here it must be noted, that when calculating the index of performance, i.e. $IP$, the difficulty of the task was adjusted appropriately based on the type of task for each subject. This was done in the calculation of the index of difficulty or $ID$ for each task, where the distance $D$ in the occasion took into account the path around the obstacle when the obstacle was present. This means that when the subject was not aware of the obstacle, their only task was to reach the target therefore distance $D$ equalled the distance between the starting position and position of the target. When an obstacle appeared on the screen in front of the subject, however, their objective transformed from a simple reaching task to obstacle avoidance, meaning the distance $D$ now equalled the shortest path around the obstacle. This was described as:

\begin{equation}
    D = 2\sqrt{\left(\frac{d}{2}\right)^2 + \left(\frac{o}{2} \right) ^2}
\end{equation}
where $d$ is the distance between the starting position and the position of the target and $o$ is the width of the obstacle the subject must avoid.

\textit{\textbf{Obstacle collision:}} Obstacle collisions were accounted for during the experiments. This was done inside the virtual environment, where scores were kept for each time the virtual dot, which was controlled by the dyad, touched the obstacle. When it did, the user interface informed the subject with the obstacle that they had failed the task and added an additional mark to the collision counter. At the same time, the partners were decoupled so that the subject who was not aware of the obstacle could finish their reaching task. The subject who was not aware of the obstacle was not informed that the other subject failed.

\textit{\textbf{Distance from obstacles:}} In the same regard, the distance between the controlled point and the obstacles was analysed in the successful trials, where there had been no collision. This was done to observe whether the distance between the virtual dot and the obstacle is influenced by the distance from the starting position to the obstacle and whether it is influenced by the dynamics within the dyad. In the latter, the aim was to observe whether there is a significant difference in the distance between the obstacle and the controlled point in dyads where the presumed leader has the obstacle and in dyads where the presumed follower has the obstacle.  

\textit{\textbf{NASA Task Load Index:}} In the analysis of the NASA-TLX questionnaire the subjects were first sorted based on whether they had an obstacle or not. This was done in order to observe how the subject's internal evaluation changed based on their knowledge of the task at hand. The scoring of the NASA-TLX was then analysed for each group as is prescribed in \cite{tlx}.

\begin{figure*}[h!]
	\centering
	\includegraphics[clip,trim=1.1cm 0cm 1.5cm 0cm,width=1\textwidth]{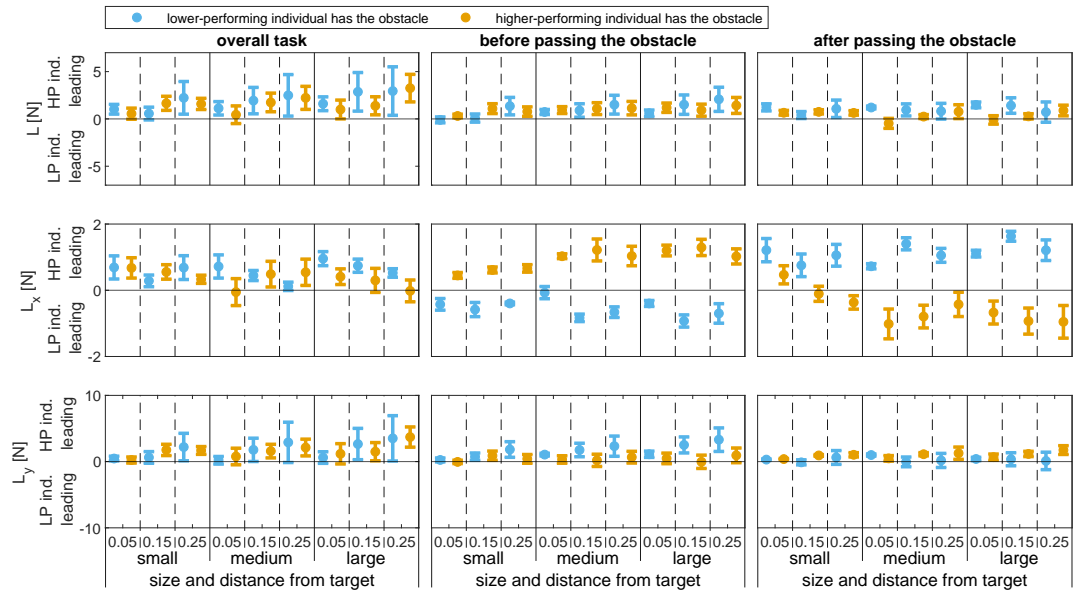}
	\caption{ Leading  subject  for  each  target in x- (middle row), y- (bottom row) and combined (top row) direction. The values in each graph represent mean leading partner L[N] ± SEM, calulated using Eq. \ref{eq:leader}, for each target used in the experiment. The far left column represents the leader of the overall task, meaning which subject was the overall leader from start until reaching the target. The middle column shows the leading partner before subjects passed the obstacle and the far right column represent the leading partner after the obstacle was passed.}\label{fig:leader}
\vspace*{-0.4cm}
\end{figure*}

\section{Results}\label{sec:results}

\begin{figure*}[h]
    \centering
    \includegraphics[clip,trim=1.75cm .5cm 1.75cm 0cm,width=1\textwidth]{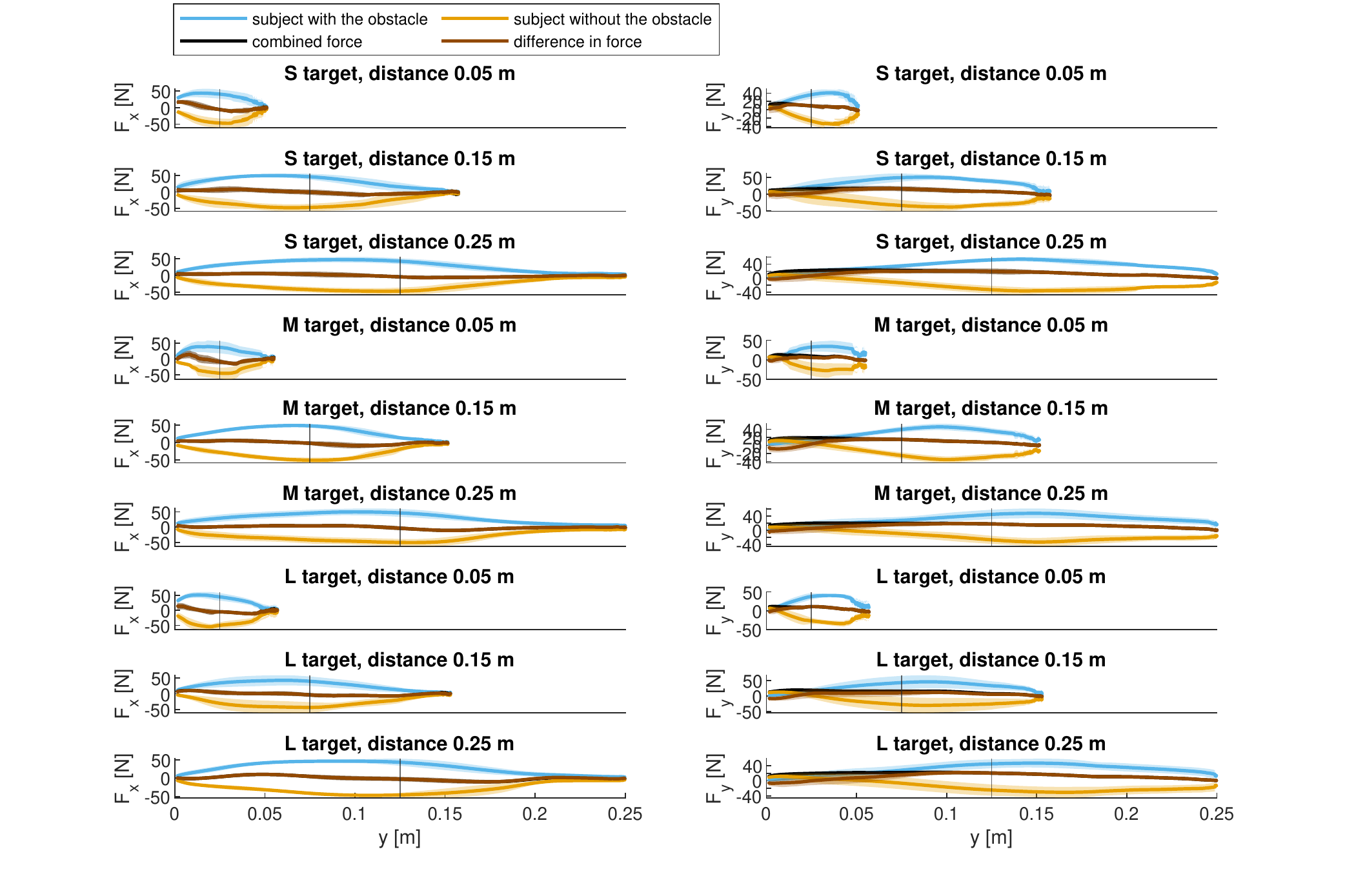}
    \caption{Average force trajectories with standard deviation (shaded area) in x- and y-direction for a typical pair during collaborative task. The figure shows force trajectories for each target type used in the experiment.}
    \label{fig:p2}
\end{figure*}

\textit{\textbf{Determining the leader-follower dynamics:}} Fig. \ref{fig:leader} shows the average leading subject in a dyad for each target in x-, y- and combined direction. The figure also shows the leader dynamic of the overall task, as well as the leader dynamic before and after passing the obstacle. Although it must be noted that ANOVA analysis has shown that neither size nor distance of the target have any significant influence on the leader dynamic. This does not coincide with the results from our previous study \cite{Kropivsek2021}. However, when separating into two parts - before and after passing the obstacle - we can see that there is a distinct switch in the leader dynamic when it comes to the x-axis. Here we can see that the subject with the obstacle assumes the role of the leader in the x-direction before the obstacle is passed to avoid it. After the obstacle is passed, however, the roles switch. This can be seen in Fig. \ref{fig:p2}, where the difference in forces between the subjects in the x-direction goes from the subject with the obstacle exerting more force before reaching the obstacle (black vertical line), while after passing the obstacle the subject with the obstacle exerts more force.

\begin{figure}[h]
	\centering
	\includegraphics[clip,trim=0.4cm 0cm 1cm 0cm,width=0.75\textwidth]{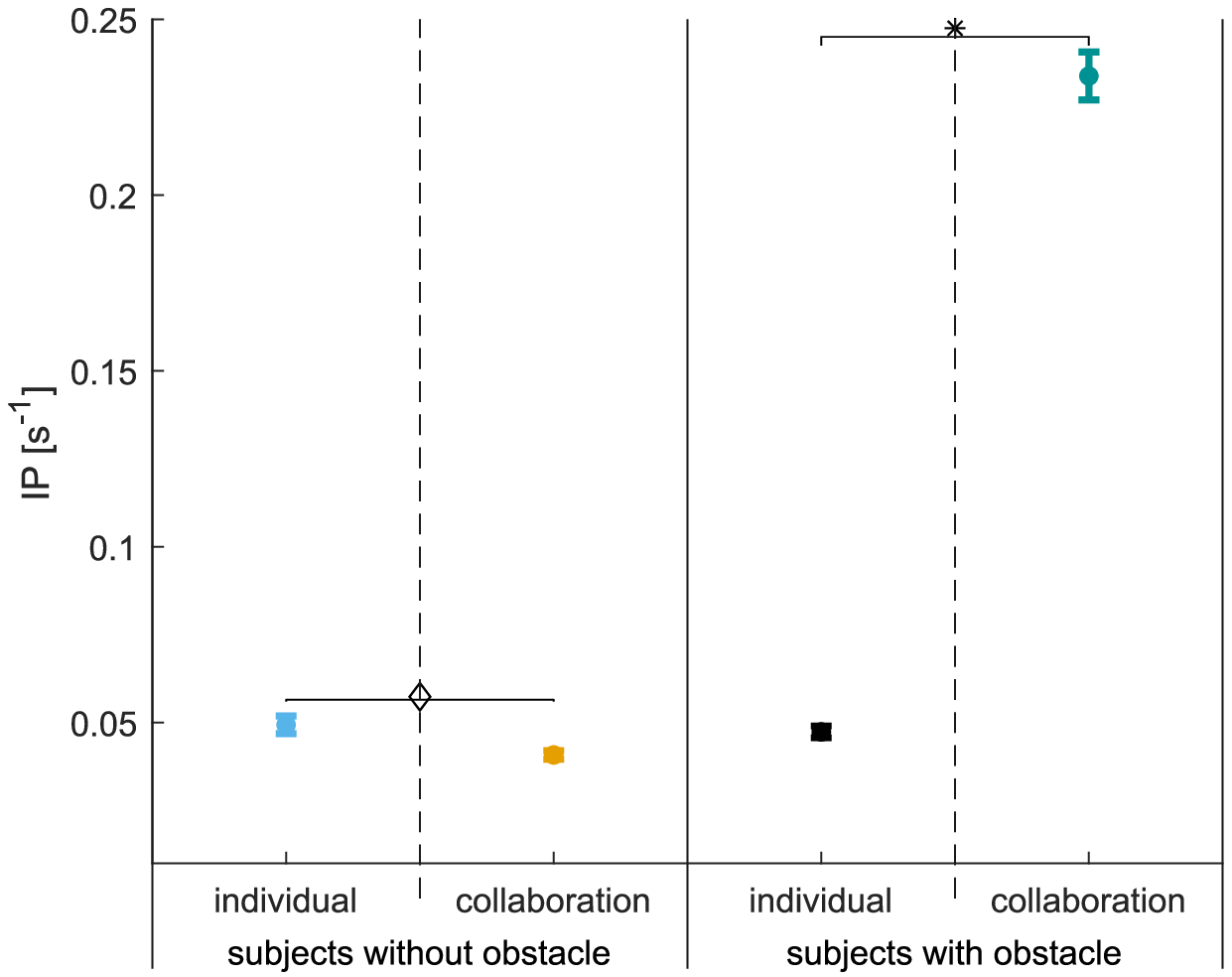}
	\caption{Task performance in individual and collaborative trials. The subject performance is shown in respect to their task - i.e. whether the subject was aware of the obstacle or not.}\label{fig:performance}
\vspace*{-0.4cm}
\end{figure}

\textit{\textbf{Task performance:}} Looking at Fig. \ref{fig:performance} we can see that the task performance decreased when subjects were collaborating compared to their individual trials for subjects who were not aware of the obstacle, while the task performance increased significantly for the subjects who were aware of the obstacle and tried to avoid it. The decrease in performance for the subjects who weren't aware of the obstacle was however not statistically significant, while the increase in the subjects who were aware of the obstacle was, with the results from post-hoc t-test showing $t_{(7)} = -32.9502$ and $p < 0.001$.

\begin{figure}[h]
	\centering
	\includegraphics[clip,trim=0cm 0cm 0cm 0cm,width=0.75\textwidth]{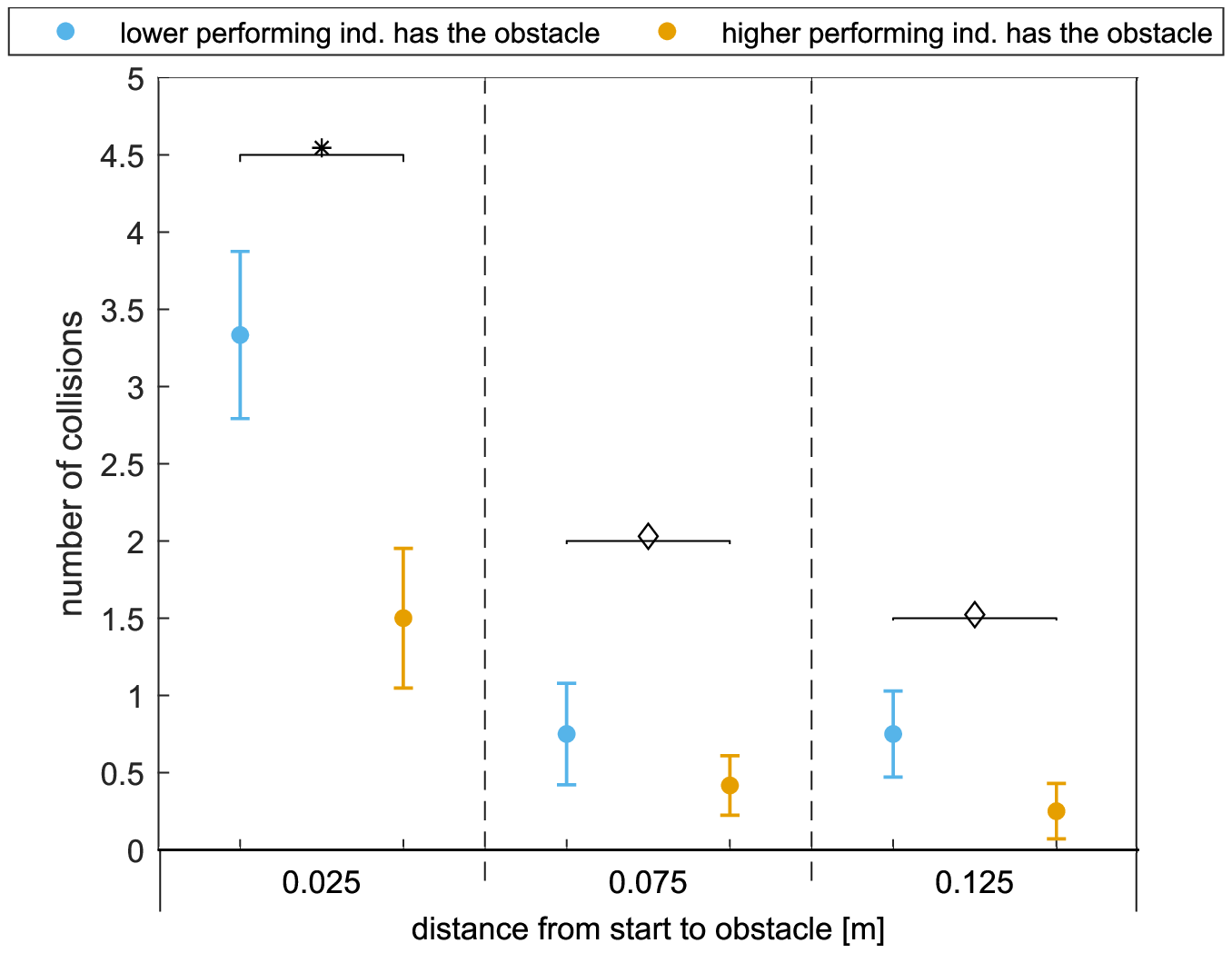}
	\caption{Average number of obstacle collisions for each group set.}\label{fig:collision}
\vspace*{-0.4cm}
\end{figure}

\textit{\textbf{Obstacle collision:}} Fig. \ref{fig:collision} presents the average number of obstacle collisions in the overall experiment. Looking at results we can notice a significant decrease in obstacle collision with the increase of the distance between the starting point and the obstacle when the presumed follower has the obstacle ($F = 16.106$, $p = 0.0039$). However looking at the dyads where the presumed leader has the obstacles we see the distance from the starting point to the obstacle has no significant influence on the resulting average number of collisions ($F = 4.776$, $p = 0.0574$). We can also see that the number of collisions is lower when the subject with the higher individual performance is the one who sees the obstacle. However, a significant difference can only be seen at the closest distance of the obstacle - i.e. at 0.025 m.

\begin{figure}[h]
	\centering
	\includegraphics[clip,trim=0cm 0.3cm 0cm 0cm,width=0.75\textwidth]{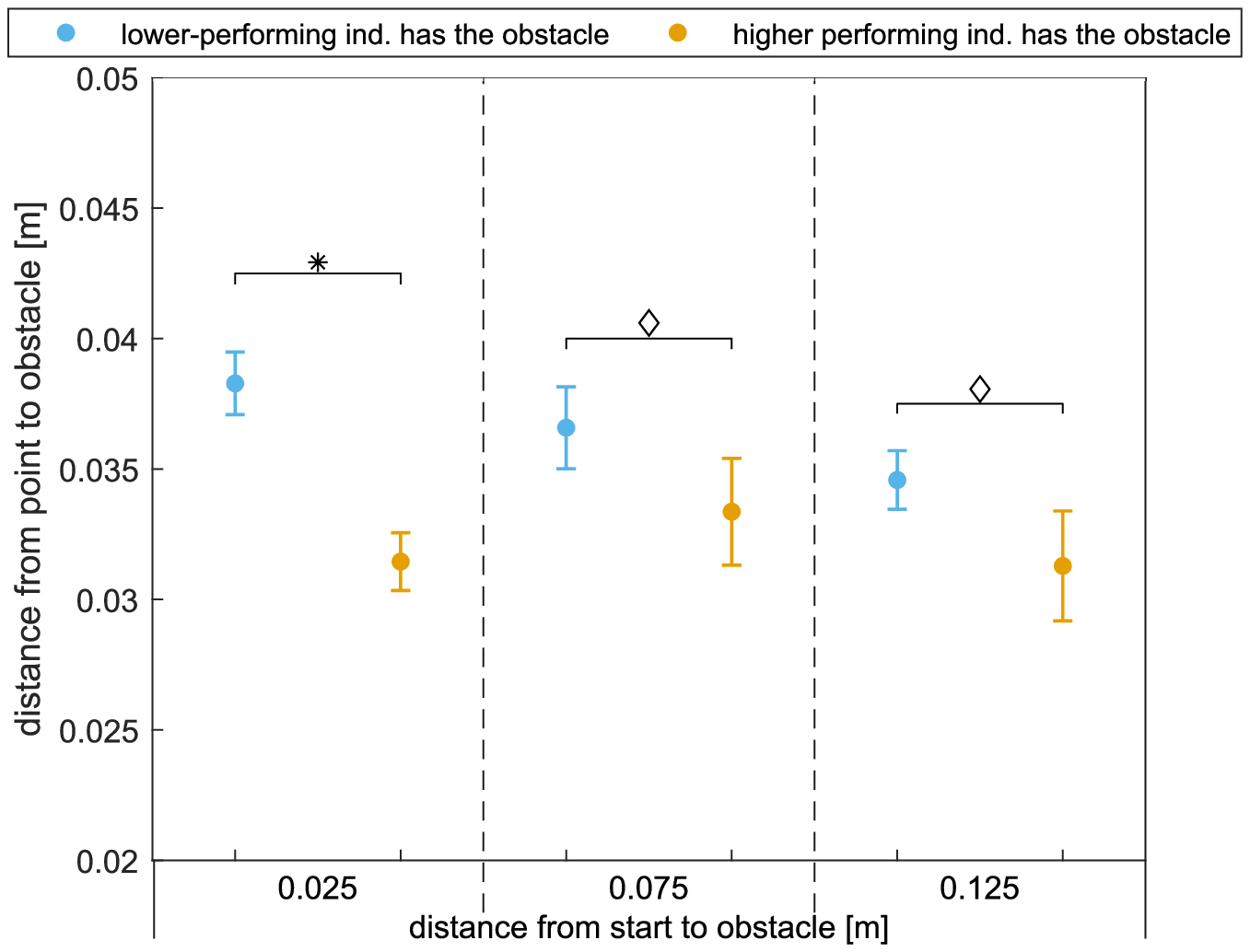}
	\caption{Average distance from obstacle in successful trials for each group set.}\label{fig:odist}
\vspace*{-0.4cm}
\end{figure}

\textit{\textbf{Distance from obstacles:}} Fig. \ref{fig:odist} clearly shows that the distance from the starting position to the obstacle has no significant influence on what the distance between the controlled point and the obstacle will be when the subjects pass the obstacle successfully. This can be confirmed both for dyads where the presumed follower had the obstacle ($F = 1.738$,  $p = 0.2539$) and for dyads where the presumed leader had the obstacle ($F = 0.706$,   $p = 0.5303$). However, we can see that in dyads where the presumed leader had the obstacle, the subjects passed the obstacle much closer than the in dyads where the presumed follower had the obstacle, although the statistical significance between the two groups has only been supported in trials where the obstacle was the closest to the starting position.

\begin{figure*}[b]
	\centering
	\includegraphics[clip,trim=1.1cm .8cm 1.7cm 0cm,width=\textwidth]{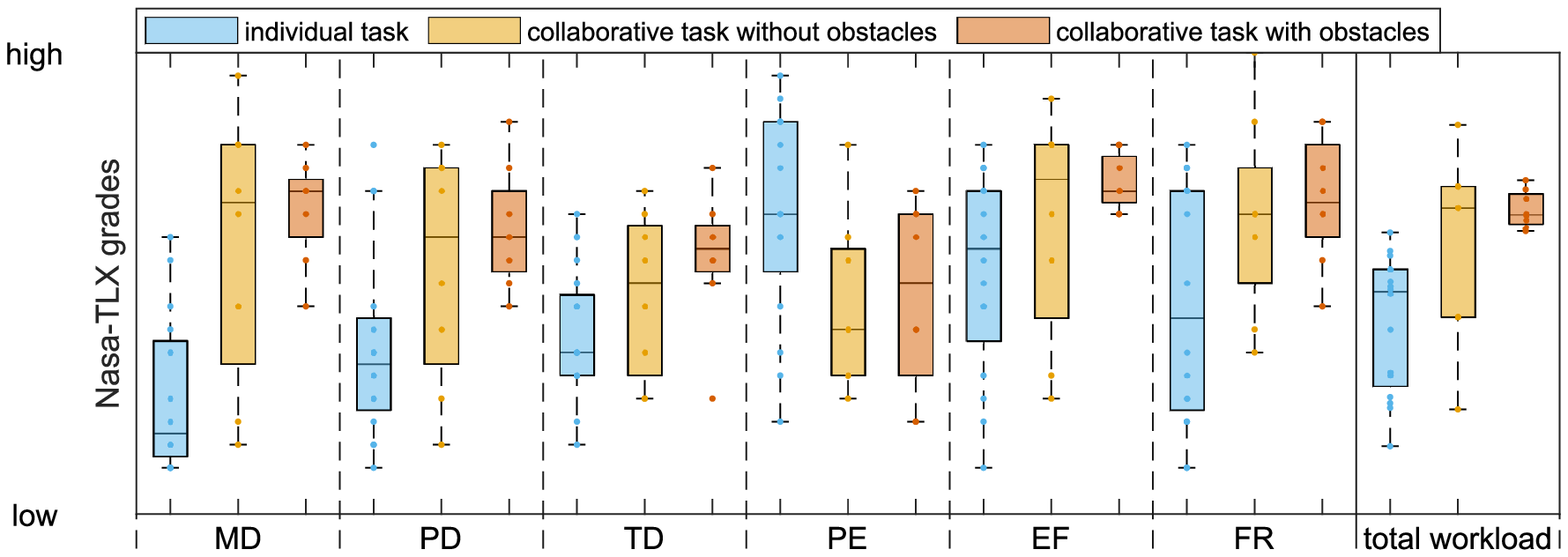}
	\caption{ NASA Task Load Index grades from all participants. The Task Load grades are presented for each factor, where MD = mental demand, PD = physical demand, TD = temporal demand, PE = performance, EF = effort, FR = frustration. The individual task includes all subject while the collaborative task grades were separated based on the subject having the obstacle or not.}\label{fig:tlx}
\vspace*{-0.4cm}
\end{figure*}

\textit{\textbf{NASA Task Load Index:}} Fig. \ref{fig:tlx} shows the subject's subjective NASA-TLX scoring. Here we can see that the subjects who had the additional task of avoiding the obstacle on average perceived the task as having a higher workload than the subjects who did not know there was an obstacle, with the scoring from the latter group being more scattered. However, we can see that in both groups the individual task was shown to have a lower workload on average.

\section{Discussion}\label{sec:discussion}

The aim of this study was to build upon the research on leader-follower dynamics previously studied in \cite{Leskovar2020}. In this paper, the main contribution was the study of the leader-follower dynamics in the additional task of obstacle avoidance as it is more complex than a simple reaching task. We hypothesised that the subject with higher-performing individual performance will still naturally assume the role of the leader, however in cases where the subject with lower individual performance has an additional task of avoiding the obstacle they will have to assume the role of the leader in order for the dyad to successfully complete the task, although the role may not be natural to them. Furthermore, we hypothesised that the dyads where the subject with higher-performing individual performance has the additional task of avoiding the obstacle will perform the task better than the dyads where the subject with lower individual performance is the one to see the obstacle. Meaning the dyads where the higher-performing individual sees the obstacle will have a lower average collision rate (in other words a higher success rate) compared to the dyads where the lower individual is the one who sees the obstacle. 

\textit{\textbf{Leader-follower dynamics:}} Results of the leader-follower dynamics seem to somewhat confirm the hypothesis set in the beginning of this paper. Although in the overall task the higher-performing individual still assumes the role of the leader regardless of whether they are aware of an obstacle or not, breaking down the calculation into x- and y- direction we can see that this is due to their dominance in forces applied to the y-direction, which is the direction the subjects had to push to reach the target. An interesting result is however seen when looking at the relationship between the forces of the subjects in the x-direction, which is the direction in which the subjects had to exert force in order to avoid the obstacle. Here we see that the leader, i.e. the partner who exerts more force over the task, is the subject that is aware of the obstacle regardless of whether they had higher or lower performance when performing the task individually. However, this can only be seen before the subjects pass the obstacle, while afterwards the roles seem to switch. We hypothesise that the switch is due to the subjects who are aware of the obstacle releasing their grip in the x-direction after successfully avoiding the obstacle, while their partner, unknowing of the additional task, continued to push with the same amount of force in the opposite direction as they were trying to correct the outside forces, keeping the path to the target as straight as possible. This can be seen as a change in the difference in forces shown in Fig. \ref{fig:p2}, where the subject who is aware of the obstacle exerts more force before they pass the obstacle, while after passing the obstacle lowers their force, allowing the other subject to become the dominant one. This change causes the role of the leader to be allocated to the subject without the obstacle.

\textit{\textbf{Task performance:}} Comparing the change in task performance of subjects who were aware of the obstacle and subjects who weren't, the results show that the performance increased significantly for the subjects who needed to avoid the obstacle, while for subjects whose objective was only to reach the target the performance, although insignificantly, decreased. These results were to be expected as for the subjects whose only goal was to reach the target, the partner they were connected to became a hindrance as they were pushing the controlled point away from the straight path towards the target. On the other hand, the performance increased for the subjects whose task was not only to reach the target but to avoid an emerging obstacle as well, as the difficulty of the task increased so significantly that although the time it took for the subjects to reach the target was longer than in individual tasks, the performance still increased. For example, comparing the largest index of difficulty or $ID$ of the reaching tasks (i.e. the $ID$ of the smallest target with the largest distance) is 0.273, while for the same target the $ID$ in an obstacle avoiding task is 0.895, implying an almost triple increase in the difficulty.

\textit{\textbf{Obstacle collision:}} Looking at the results in Fig. \ref{fig:collision}, we can see that they confirm our established hypothesis, as we can see that the dyads where the subject with higher individual performance sees the obstacle have a lower average collision rate in an experiment than dyads where the subject with lower individual performance sees the obstacle. This can be explained through the results from the study of leader-follower dynamics in our previous paper \cite{Kropivsek2021}, where we showed that the subject with higher individual performance naturally assumes the role of the leader. Due to this, they have better control over the collaborative task and thus have an easier job of controlling the point away from the obstacle.

However, we can see that there is a significant difference in the number of collisions only in obstacles that are closest to the starting position. From these results, we can gather that in both experiment groups subjects who saw the obstacles were able to adjust the movement to avoid the obstacle when it was further away, while when the target was closest to the starting position the success rate depended mostly on the subjects' reaction time and force applied to the x-axis.

Furthermore, we can see that the number of collisions decreases with distance from the obstacle for both groups of subjects. From this, we can see that tasks where the targets, and hence the obstacles, are closer to the starting position were indeed more difficult than those where the distance to the target is greater. This is in direct contradiction to the definition of the performance index of Fitts' law \cite{Fitts1954}, according to which greater distances lead to higher task difficulty. Thus, we must conclude that the index of performance (IP) used to calculate objective task performance is not fully applicable in our case.

\textit{\textbf{Distance from obstacles:}} Figure \ref{fig:odist} shows that the distance between the controlled point and the obstacle when the subjects were passing it remains similar regardless of the obstacle distance from the start. However, the results imply that when the lower-performing individual, who is presumed to be the follower, is the subject aware of the obstacle, the distance from the obstacle is larger than when the partner who sees the obstacle is the subject with higher individual performance, i.e. the presumed leader. Here we can hypothesise that this might be due to the imbalance in forces applied by both partners throughout the overall task. Namely, in order to avoid the obstacle, the subjects with lower individual performance exaggerated their movement around the obstacle as they felt less confident and comfortable in avoiding the obstacle due to the fact that they had less control over the movement of the controlled point on the screen. This is however only our interpretation of the results and cannot be proven.

\textit{\textbf{NASA Task Load Index:}} Results of the NASA-TLX inquiry showed that both subjects who are and those who aren't aware of the obstacle, experienced the collaborative task as more frustrating and having a higher workload. These results seem to be contrary to our previous study in \cite{Kropivsek2021icar}, where the subjects rated the collaborative task with a human the second least frustrating and second easiest right after the human-robot collaborative task. However, the results seen in the current study are to be expected as the experiment is not exactly the same as in our previous ones. Here, we introduced the obstacle avoidance task from which the recorded frustration seems to stem from. As the obstacle is introduced to only one partner and not both, the interpretation of the task at hand was different for the two subjects in a dyad. Due to this the partner who wasn't aware of the obstacle understood their partner's force away from the straight path to the target as a hindrance, as this obstructed them from following the shortest path towards the target. On the other hand, the partner who was aware of the obstacle understood the force of their partner who was trying to correct their force to stay on the straight line as pushing them into the obstacle, also resulting in a sense of hindrance. 

The mere addition of the obstacle avoidance task should not be seen however as the main cause for the participants' frustration. Instead, we presume that the frustration of all subjects involved is rooted in the lack of communication between the partners that would allow them to perform their objectives efficiently. This shows that the current experimental setup does not allow the subject to communicate with each other effectively enough to be able to understand the full picture of the task or in other words -- a collaboration between the partners can only be efficient when they are able to determine a mutual objective, which in this experiment was not the case. These experiment types are important to observe as in real-world scenarios, not both collaborating partners have the same perception of the task at hand, as was also emphasised in \cite{Grosz1996}\cite{Cohen1991}.

\textit{\textbf{Contributions:}} Results presented in this study may prove to be beneficial in better understanding how people collaborate in more complex environments, where the task is not as clear as it seems at first glance. Furthermore, results from human collaboration studies are always beneficial in the development of novel robot control models as it allows researchers to develop more human-like interactions between the human and robot partner based on peer-reviewed evidence. As stated in the introduction, implementing human-like interaction dynamics into HRC aids in the further improvement of HRC as studies such as \cite{Ganesh2014}\cite{Noohi2016}\cite{Ivanova2020}  showed that the human partners find human-like behaviour of robots more intuitive. Implementing the leader-follower role allocation studied in this paper into a robot control model could allow the robot partner to assume either the role of a follower or that of a leader when necessary, making not only the robot more human-like but the whole interaction. 

\section*{Declarations}

\subsection*{Funding}
This work was supported by Slovenian Research Agency grant N2-0130.

\subsection*{Conflict of interest}
The authors have no relevant financial or non-financial interests to disclose.

\subsection*{Ethics approval} 
This study was conducted in accordance with the code for ethical conduct in research at Jožef Stefan Institute (JSI) and was approved by the National Medical Ethics Committee (No.: 0120-228/2020-3, approved on 13.7.2020).

\subsection*{Consent to participate}
Prior to conducting the experiment, all participants were informed about the experimental procedure, potential risks, the aim of the study and gave their written informed consent in accordance with the code for ethical conduct in research at Jožef Stefan Institute (JSI).

\subsection*{Consent for publication}
All authors have read and agreed to the publication of the final version of the manuscript.

\subsection*{Availability of data, materials and code}
All materials, data and code used in this study are available upon request to the authors.

\bibliography{main}

\end{document}